\documentclass[journal]{IEEEtran}
\usepackage[pdftex]{graphicx}
\usepackage{color}
\usepackage{url}
\usepackage{tikz}
\usepackage{framed}
\usepackage{cite}
\usepackage{ifthen}

\usepackage{amsmath}
\usepackage{amssymb}
\usepackage{amsfonts}
\usepackage{bm}
\usepackage{amsthm}
\usepackage{mathtools}
\usepackage{latexsym} 

\usepackage{algorithm}
\usepackage{algorithmic}
\usepackage{booktabs}
\usepackage{multirow}
\usepackage{enumerate}
\usepackage{siunitx}
\usepackage[caption=false,font=footnotesize]{subfig}
\usepackage{array}

\newtheorem{theorem}{Theorem}
\newtheorem{remark}{Remark}
\newtheorem{prop}{Proposition}

\newcommand{\paren}[1]{\left(#1\right)}

\newcommand{\E}{\mathrm{e}}

\DeclareMathOperator{\diag}{diag}
\DeclareMathOperator{\tr}{tr}

\newcommand{\INLINEIF}[2]{\ifthenelse{#1}{#2}{}}

\renewcommand{\Re}{\operatorname{Re}}

\renewcommand{\Im}{\operatorname{Im}}

\mathtoolsset{showonlyrefs=true}  % 参照していない式番号を表示しない

\newcommand{\tblack}{\textcolor{black}}

\DeclareUnicodeCharacter{221E}{\ensuremath{\infty}}

\begin{document}
\title{Compression Method for Deep Diagonal State Space Model Based on $\mathcal{H}^{2}$ Optimal Reduction}
\author{Hiroki Sakamoto and Kazuhiro Sato\thanks{H. Sakamoto and K. Sato are with the Department of Mathematical Informatics, Graduate School of Information Science and Technology, The University of Tokyo, Tokyo 113-8656, Japan, email: soccer-books0329@g.ecc.u-tokyo.ac.jp (H. Sakamoto), kazuhiro@mist.i.u-tokyo.ac.jp (K. Sato) }}

\maketitle
\thispagestyle{empty}
\pagestyle{empty}

\begingroup
\renewcommand\thefootnote{}%  番号を消す
\footnotetext{%
Accepted to \emph{IEEE Control Systems Letters}.}
\endgroup

\begin{abstract}
Deep learning models incorporating linear SSMs have gained attention for capturing long-range dependencies in sequential data.
However, their large parameter sizes pose challenges for deployment on resource-constrained devices.
In this study, we propose an efficient parameter reduction method for these models by applying $\mathcal{H}^{2}$ model order reduction techniques from control theory to their linear SSM components.
In experiments, the LRA benchmark results show that the model compression based on our proposed method outperforms an existing method using the Balanced Truncation, while successfully reducing the number of parameters in the SSMs to $1/32$ without sacrificing the performance of the original models.
\end{abstract}


\IEEEpeerreviewmaketitle

\section{Introduction} \label{sec:intro}

% DDSSMsについて
Deep learning models that incorporate linear State Space Models (SSMs)~\cite{kalman1960new} have achieved remarkable success across various fields, including text~\cite{mehta2023long}, audio~\cite{jiang2025dual}, images~\cite{ma2024u}, and videos~\cite{li2024videomamba}. Since the introduction of this class of models in works such as~\cite{gu2020hippo, gu2021combining, gu2022efficiently}, these architectures have attracted increasing attention, with several studies demonstrating their potential to model long-range dependencies in sequential data~\cite{gu2022efficiently, gupta2022diagonal, gu2022parameterization, smith2023simplified, gu2023mamba}.
A notable direction in this research is the imposition of diagonal constraints on the internal linear SSMs, enabling stable and computationally efficient training procedures~\cite{gupta2022diagonal, gu2022parameterization, nguyen2022s4nd, gu2023mamba}. In this study, we refer to deep learning models that leverage such diagonal state-space structures as Deep Diagonal State Space Models (DDSSMs).

% MOR-based DDSSMsの圧縮
While DDSSMs exhibit high modeling capability, the number of parameters increases significantly when the state dimension $N$ becomes large, which poses challenges for practical applications. 
To address this issue, recent studies~\cite{ezoe2024model, forgione2024model, gwak2024layer} have explored compression techniques for DDSSMs based on Model Order Reduction (MOR), a methodology for constructing Reduced-Order Models (ROMs) by reducing the state dimension of linear SSMs.
In particular,~\cite{ezoe2024model, forgione2024model} utilize the infinite-time Balanced Truncation (BT) method~\cite{moore1981principal, antoulas2005approximation}, while~\cite{gwak2024layer} employs a metric based on the $\mathcal{H}^\infty$ norm, referred to as the LAST score.
In both approaches, the state dimension $N$ of diagonal linear SSMs is reduced to achieve a smaller number of parameters.
However, these $\mathcal{H}^\infty$-based MOR methods do not guarantee any optimality in the context of model reduction. 
Moreover, although the length of input-output sequence data is inherently finite in practice, these methods implicitly assume infinite-time behavior of the original SSMs for the purpose of model reduction.

% 線形SSMに対する H2 MOR のサーベイ
In contrast to model reduction methods that focus on the $\mathcal{H}^\infty$-norm, such as BT method, the framework of $\mathcal{H}^2$-based MOR~\cite{antoulas2005approximation, goyal2019time, sinani2019h2, das2022h, sakamoto2024data} enables the development of algorithms with provable optimality guarantees. However, existing $\mathcal{H}^2$-MOR techniques cannot be directly applied to the linear SSMs appearing in DDSSMs due to their specific properties.

% 本研究の目的と貢献
Motivated by the aforementioned challenges, the primary goal of this study is to develop an efficient compression method for DDSSMs based on the $\mathcal{H}^{2}$-MOR framework. To this end, we propose an $\mathcal{H}^{2}$-based MOR tailored to the linear SSMs in DDSSMs that exhibit unique structural characteristics. Our contributions can be summarized in two key points:

\begin{enumerate}
    \item \textbf{A novel $\mathcal{H}^{2}$-MOR technique for linear SSMs in DDSSMs:}\\
          We propose a new $\mathcal{H}^2$-MOR technique that preserves the key properties of the original linear SSMs—namely, complex-valuedness, finite-time nature, diagonal structure, and stability. In particular, we formulate a gradient-based optimization algorithm for this purpose.

    \item \textbf{$\mathcal{H}^{2}$-MOR based compression methods:}\\
          We integrate the proposed $\mathcal{H}^2$-MOR technique with the model compression approach introduced in~\cite{ezoe2024model}. Even in cases where BT-based compression~\cite{ezoe2024model} fails to achieve sufficient performance, our method enables the construction of compressed DDSSMs without significant loss in accuracy. We demonstrate this on several tasks from the Long Range Arena (LRA) benchmark~\cite{tay2021long}.
\end{enumerate}

% 本論文の構成
This paper is organized as follows. 
In Section~\ref{sec:background}, after introducing the properties of the SSMs used in this study and the DDSSMs, we show a model compression method based on the BT framework and its limitation.
Section~\ref{sec:main} formulates the finite-time structure-preserving $\mathcal{H}^{2}$-MOR problem and presents a gradient-based algorithm. In Section \ref{sec:experiments}, we demonstrate that the proposed model compression approach, grounded in the new $\mathcal{H}^{2}$-MOR technique, achieves favorable accuracy on LRA benchmark tasks. Finally, we conclude the paper with a summary in Section \ref{sec:conclusion}.

\textbf{Notation.} 
\tblack{\(\|A\|\) denotes the 2-norm when $A$ is a vector and the Frobenius norm when $A$ is a matrix.}
For \(A \in \mathbb{C}^{N \times N}\), \(A^\top\) and \(A^*\) denote the transpose and Hermitian transpose, respectively.  
\(\mathrm{i}\) is the imaginary unit.  
\(\diag(\Lambda)\) denotes the diagonal matrix with entries \(\Lambda \in \mathbb{C}^N\).  
For \(\alpha \in \mathbb{C}\), \(\mathrm{Re}(\alpha)\) and \(\mathrm{Im}(\alpha)\) are its real and imaginary parts.  
For \(a, b \in \mathbb{C}^N\), \(a \odot b\) denotes the Hadamard product.  
For a smooth \(f:\mathbb{R}^n\to\mathbb{R}\), \(\nabla f(x)\) denotes its gradient.  
For a smooth \(f:\mathbb{C}^n\to\mathbb{R}\), we define \(\nabla_x f := \nabla_{\Re(x)} f + \mathrm{i} \nabla_{\Im(x)} f\).  
For \(x \in \mathbb{R}^N\), \(\exp(x)\) denotes the element-wise exponential: \((\exp(x))_i = \exp(x_i)\).  
\(\mathbb{C}_{-}^N := \{ z \in \mathbb{C}^N \mid \mathrm{Re}(z_i) < 0 \ \forall i \}\) is the set of vectors with strictly negative real parts.  

\section{Preliminaries} \label{sec:background}

% 本節の概要
We first introduce the Diagonal State Space (DSS) model, followed by the deep models (DDSSMs) that incorporate it. We then review the BT-based compression method for DDSSMs~\cite{ezoe2024model} and its limitations.

\subsection{Diagonal State Space Model} \label{subsec:ssm}

% 本研究で対象とする状態空間モデル
Consider the following SSM
\begin{align} \label{eq:c-SSM}
   \begin{cases}
    \dot{x}(t)=Ax(t) +Bu(t), \quad x(0)=0, \\
    y(t)=Cx(t), \quad t \geq 0,  
   \end{cases}
\end{align}
where $A\in\mathbb{C}^{N\times N}$ is stable, i.e., all of its eigenvalues lie in the open left-half complex plane, and $B\in \mathbb{C}^{N\times m}$, $C\in \mathbb{C}^{p\times N}$.
We refer to \eqref{eq:c-SSM} as a DSS model when $A := \mathrm{diag}(\Lambda) \in \mathbb{C}^{N \times N}$ with $\Lambda \in \mathbb{C}^{N}$.
Generally, $u(t)\in \mathbb{C}^{m}$, $y(t)\in \mathbb{C}^{p}$, and $x(t)\in \mathbb{C}^N$ denote the input, output, and state at time $t$, respectively.
The transfer function $G(s)=C(sI-A)^{-1}B\in \mathbb{C}^{p\times m}$ of the system \eqref{eq:c-SSM} is defined as the relation between the output response and the input signal in the frequency-domain with zero initial condition.

% SSMに対するH2,tauノルム
The finite-time $\mathcal{H}^{2}$-norm $\|G\|_{\mathcal{H}^{2}, \tau}$ of system \eqref{eq:c-SSM} over a limited time interval $[0, \tau]$ with $\tau < \infty$ is defined as follows:
\begin{align*}
    \|G\|_{\mathcal{H}^{2}, \tau}^2 
    &:= \int_{0}^{\tau} \tr (B^{*}\E^{A^{*}t}C^{*}C\E^{At}B)dt \\
    &= \tr (B^{*}Q_{\tau}B) = \tr (CP_{\tau}C^{*}),
\end{align*}
where the finite-time Gramians $P_{\tau}$ and $Q_{\tau}$ of system \eqref{eq:c-SSM} are defined as
\begin{align}
    P_{\tau} = \int _{0}^{\tau}\E^{At}BB^{*}\E^{A^{*}t}dt, \quad Q_{\tau} = \int _{0}^{\tau}\E^{A^{*}t}C^{*}C\E^{At}dt.
\end{align}
Here, as $\tau \to \infty$, the norm $\|\cdot\|_{\mathcal{H}^2, \tau}$ yields the infinite-time $\mathcal{H}^2$-norm $\|\cdot\|_{\mathcal{H}^2}$.
Unlike the norm $\|\cdot\|_{\mathcal{H}^{2}}$, $\|\cdot\|_{\mathcal{H}^{2}, \tau}$ is well-defined even for unstable systems, since it is computed over a finite-time horizon \cite{das2022h}.
Note that the condition $\lambda_{i}+\lambda_j^{*} \neq 0$ for all eigenvalues $\lambda_i$, $\lambda_j$ of $A$ is equivalent to that $P_{\tau}$ and $Q_{\tau}$ are the unique solutions of the following Lyapunov equations:
\begin{align}
    AP_{\tau} + P_{\tau}A^{*} + BB^{*} - \E^{A\tau}BB^{*}\E^{A^{*}\tau} 
    &= 0, \label{eq:fLyap1}\\
    A^{*}Q_{\tau} + Q_{\tau}A + C^{*}C - \E^{A^{*}\tau}C^{*}C\E^{A\tau} 
    &= 0.\label{eq:fLyap2}
\end{align}
See \cite{antoulas2005approximation, gawronski1990model} for more details.

% 出力誤差と伝達関数誤差の関係
There exists a close relationship between the error-norm of the output $y$ and the finite-time $\mathcal{H}^{2}$-norm of the transfer function $G$. 
In fact, consider a surrogate model of \eqref{eq:c-SSM}: 
\begin{align}
\begin{cases}
    \dot{\hat{x}}(t)=\hat{A}\hat{x}(t)+\hat{B}u(t),\\
    \hat{y}(t)=\hat{C}\hat{x}(t), 
\end{cases}\label{eq:r-dss}
\end{align}
where $\hat{A}\in\mathbb{C}^{r\times r},\hat{B}\in\mathbb{C}^{r \times m},\hat{C}\in\mathbb{C}^{p \times r}$.
Let $\hat{G}$ denote the transfer function of \eqref{eq:r-dss}. 
Then, the inequality holds~\cite{goyal2019time}:
\begin{align}
    \max_{t\in [0,\tau]} \|y(t)-\hat{y}(t)\| \leq \|G-\hat{G}\|_{\mathcal{H}^{2},\tau}\cdot \sqrt{\int_{0}^{\tau}\|u(t)\|^{2}dt}, \label{eq:relation-g-y}
\end{align}
which implies that if $\|G-\hat{G}\|_{\mathcal{H}^{2},\tau}$ is sufficiently small, then the outputs $y$ and $\hat{y}$ are close, provided that the input energy $\int_0^\tau \|u(t)\|^2 dt$ is small.

\subsection{Deep Diagonal State Space Models}

% DDSSMsの定義
This study focuses on deep learning models incorporating the DSS model \eqref{eq:c-SSM}, referred to as DDSSMs, including DSS~\cite{gupta2022diagonal}, S4D~\cite{gu2022parameterization}, S4ND~\cite{nguyen2022s4nd}, and S5~\cite{smith2023simplified}.  
In the numerical experiments, we use DSS~\cite{gupta2022diagonal}; see Section~\ref{sec:experiments} for details.

% DDSSMs内のDSS model
In DDSSMs, a discretized version of~\eqref{eq:c-SSM} with sampling time \(\Delta \in \mathbb{R}_{>0}\) is used for numerical computations:
\begin{align}\label{eq:d-SSM}
    \begin{cases}
    x_{k} = \bar{A}\,x_{k-1} + \bar{B}\,u_k, \quad k = 1, 2, \dots \\
    y_k = \bar{C}\,x_k,
    \end{cases}
\end{align}
where \(\bar{A} = \E^{A\Delta}\), \(\bar{B} = (\bar{A}-I)\,A^{-1}B\), \(\bar{C} = C\).
% , and \(L\) denotes the input length.  
DDSSMs are trained using optimization algorithms such as Adam, where \((A, B, C)\) in~\eqref{eq:c-SSM} are optimized to minimize a loss function.  
The matrix \(A \in \mathbb{C}^{N\times N}\) is diagonal, enabling efficient computation via the Fast Fourier Transform (FFT)~\cite{cormen2022introduction}, and assumed to be stable to keep \(\|y_k\|\) bounded.  
Under these assumptions, \(\bar{A}\) inherits diagonal structure and discrete-time stability.  
% Since the input length \(L\) is finite, we interpret the system described by~\eqref{eq:c-SSM} as operating over a finite-time horizon.
% Moreover, when the input sequence $( u_k )$ is of finite length, the system described by~\eqref{eq:c-SSM} can be interpreted as operating on a finite-time horizon.

% DDSSMs内のDSS model以外の部分
% In addition to the state-space module, DDSSMs incorporate nonlinear connections
% % , such as activation functions, 
% to capture complex temporal dynamics.  
% They also employ linear combination mechanisms to aggregate information across feature channels, enabling the modeling of inter-series dependencies that cannot be expressed by the linear SSM alone.
% Note that S5\cite{smith2023simplified} naturally incorporates such linear combination mechanisms through its use of a MIMO structure.
\tblack{In addition to the linear state-space module, a DDSSM comprises nonlinear connections (e.g., gating or activation) and a linear channel-mixing part that blends the \(H\) feature channels of each input/output vector, enabling the model to capture inter-series dependencies that a purely linear SSM cannot represent.  
In DSS, S4D, and S4ND, this channel mixing is explicitly implemented by learnable weight matrices placed before and/or after their diagonal SISO systems ($m=p=1$ for \eqref{eq:c-SSM}).  
By contrast, S5 parameterizes the SSM itself as a MIMO system ($m=p=H$ for \eqref{eq:c-SSM}); this internal structure already performs the required channel mixing, thus no separate mixing part is needed.}

\subsection{Infinite-Time Balanced Truncation-Based Compression for DDSSMs and its limitations} 
% 導入
Recent work~\cite{ezoe2024model, forgione2024model} compresses DDSSMs by applying the infinite‑time BT method, thereby lowering inference cost for step‑by‑step processing.  
This paper specifically focus on the approach of~\cite{ezoe2024model}.

% 平衡実現打ち切り法ベースのモデル圧縮法
As described in Section~\ref{subsec:compression}, \cite{ezoe2024model} applies the BT method to pre-trained DSS models \eqref{eq:c-SSM} to obtain ROMs, which are then re-trained using the obtained ROMs as initial points to construct compressed models.
Although this BT-based approach is shown to be effective for model compression, there remain several aspects that could be improved.

% BTベースのモデル圧縮の改善点1
First, in some cases, the frequency responses of the original pre-trained DSS models are not sufficiently approximated by the BT method. In such cases, the original model’s output $y$ may not accurately approximated from \eqref{eq:relation-g-y}. 

% BTベースのモデル圧縮の改善点2
Second, although the input sequence is finite, \cite{ezoe2024model} employs an infinite-time BT method. 
Consequently, it approximates an infinite-time SSM, despite the original target being a finite-time SSM.
Note that the finite-time BT method \cite{gawronski1990model, kurschner2018balanced} may yield unstable ROMs, leading to compression failures (Section~\ref{sec:experiments}).

% BTベースのモデル圧縮の改善点まとめ
The performance of DDSSMs depends on the initial SSMs used for training~\cite{gu2020hippo}. Therefore, if the MOR is not performed effectively, the re-trained model may not achieve sufficient performance.

%%%%%%%%%%%%%%%%
\section{Finite-Time $\mathcal{H}^{2}$-MOR Preserving Complex Diagonal Structure}\label{sec:main}
% 導入
To enable model compression of DDSSMs, we introduce an $\mathcal{H}^{2}$-MOR tailored to the DSS model~\eqref{eq:c-SSM}, which exhibits the distinctive properties found in DDSSMs.
We first formulate the $\mathcal{H}^{2}$-MOR problem as an optimization problem for a general class of SSMs, including MIMO systems, and then propose a gradient-based algorithm to solve it.

\subsection{Finite-time $\mathcal{H}^{2}$ Model Order Reduction Problem}\label{subsec:mor-problem}

% 望まれる構造
As described in Section~\ref{sec:background}, DSS models \eqref{eq:c-SSM} in DDSSMs are designed to satisfy the following specialized properties: (i) the matrix $A$ is diagonal and stable, (ii) the parameters are complex-valued, and (iii) the system operates over a finite-time horizon, determined by the finite length of the input sequence.

% 最適化問題
Motivated by the relation \eqref{eq:relation-g-y}, and while preserving these properties, we consider the finite-time $\mathcal{H}^{2}$-MOR problem, which is formulated as
\begin{align}
\label{eq:opt_original}
\begin{aligned}
    &\text{minimize} &&\|G - \hat{G}\|_{\mathcal{H}^{2},\tau}^2 \\
    &\text{subject to} &&(\hat{\Lambda}, \hat{B}, \hat{C})\in \mathbb{C}_{-}^{r}\times \mathbb{C}^{r\times m}\times \mathbb{C}^{p\times r},
\end{aligned}
\end{align}
where $G$ and $\hat{G}$ denote the transfer functions of the original system \eqref{eq:c-SSM} with $A:=\diag(\Lambda)\in\mathbb{C}^{N\times N}$ and the ROM \eqref{eq:r-dss} with $r\ll N$ and $\hat{A}:=\diag(\hat{\Lambda})\in\mathbb{C}^{r\times r}$, respectively.
The first-order necessary optimality conditions for \eqref{eq:opt_original} have been derived in \cite{goyal2019time, sinani2019h2}.

% 目的関数の書き換え
\begin{prop}\label{prop:modified}
    The objective function $\|G - \hat{G}\|_{\mathcal{H}^{2},\tau}^2$ of \eqref{eq:opt_original} are rewritten as
    \begin{align*}
        \|G - \hat{G}\|_{\mathcal{H}^{2},\tau}^{2} &=  \tr (B^{*}Q_{\tau}B) + f(\hat{\Lambda}, \hat{B}, \hat{C}) \\
        &=  \tr (CP_{\tau}C^{*}) + f(\hat{\Lambda}, \hat{B}, \hat{C}).
    \end{align*}
    Here,
    \begin{align}
        f(\hat{\Lambda}, \hat{B}, \hat{C})
        &= \tr (\hat{B}^{*}\hat{Q}_{\tau}\hat{B} + 2\Re(\hat{B}^{*}Y_{\tau}^{*}B)) \\
        &= \tr (\hat{C}\hat{P}_{\tau}\hat{C}^{*} - 2\Re(\hat{C}X_{\tau}^{*}C^{*})),\label{eq:mod_obj}
    \end{align}
    and $P_{\tau}$ and $Q_{\tau}$ satisfy \eqref{eq:fLyap1} and \eqref{eq:fLyap2}, respectively, and $X_{\tau}$, $Y_{\tau}$, $\hat{P}_{\tau}$, $\hat{Q}_{\tau}$ satisfy the following equations with $A:=\diag(\Lambda)$ and $\hat{A}:=\diag(\hat{\Lambda})$:
    \begin{align}
        \hat{A}\hat{P}_{\tau} + \hat{P}_{\tau}\hat{A}^{*} + \hat{B}\hat{B}^{*} - \E^{\hat{A}\tau}\hat{B}\hat{B}^{*}{\E^{\hat{A}^{*}\tau}} &= 0, \label{eq:f-lyap1}\\
        \hat{A}^{*}\hat{Q}_{\tau} + \hat{Q}_{\tau}\hat{A} + \hat{C}^{*}\hat{C} - \E^{\hat{A}^{*}\tau}\hat{C}^{*}\hat{C}{\E^{\hat{A}\tau}} &= 0, \label{eq:f-lyap2}\\
        AX_{\tau} + X_{\tau}\hat{A}^{*} + B\hat{B}^{*} - \E^{A\tau}B\hat{B}^{*}{\E^{\hat{A}^{*}\tau}} &= 0, \label{eq:f-sylve1}\\
        A^{*}Y_{\tau} + Y_{\tau}\hat{A} - C^{*}\hat{C} + \E^{A^{*}\tau}C^{*}\hat{C}{\E^{\hat{A}\tau}} &= 0, \label{eq:f-sylve2}
    \end{align}
    where \eqref{eq:f-lyap1} and \eqref{eq:f-lyap2} are the Lyapunov equations and \eqref{eq:f-sylve1} and \eqref{eq:f-sylve2} are the Sylvester equations.    
\end{prop}
\begin{proof}
    \tblack{The proof follows from \cite[Proposition~2.2]{goyal2019time} by extending the setting to complex matrices and restricting the $A$ matrix to be diagonal.
    \tblack{See Appendix~\ref{app:proof_of_prop1}.}
    }
\end{proof}

% 修正した最適化問題
Consequently, the following equivalent optimization problem for \eqref{eq:opt_original} is obtained from Proposition \ref{prop:modified}:
\begin{align}
\label{eq:opt_modified}
\begin{aligned}
    &\text{minimize} && f(\hat{\Lambda}, \hat{B}, \hat{C}) \\
    &\text{subject to} && (\hat{\Lambda}, \hat{B}, \hat{C}) \in \mathbb{C}_{-}^{r} \times \mathbb{C}^{r\times m} \times \mathbb{C}^{p\times r}.
\end{aligned}
\end{align}
Solving~\eqref{eq:opt_modified} yields a ROM that approximates the original DSS model~\eqref{eq:c-SSM} in the finite-time \(\mathcal{H}^2\)-norm while preserving its properties.  
As implied by~\eqref{eq:relation-g-y}, this corresponds to generating a lower-parameter DSS model in DDSSMs that approximates the output \(y\) for the same input \(u\).
\tblack{Note that, since the infinite-time BT used in \cite{ezoe2024model} and the proposed method construct ROMs under different evaluation metrics, a direct comparison of their input--output behavior is difficult.}
\tblack{See Appendix~\ref{app:evaluation_metrics} for details about the evaluation metrics.}

\subsection{Gradients for complex variables}\label{subsec:gradients}

% 修正した最適化問題に対する勾配
The optimization problem \eqref{eq:opt_modified} is nonconvex. 
We derive the gradients for complex variables to construct the gradient-based algorithm for \eqref{eq:opt_modified}. 
Because the optimization variables are complex, we decompose them into real and imaginary parts.
\begin{theorem}
\label{thm:gradients}
    For the objective function of  \eqref{eq:opt_modified}, the gradients $\nabla_{\hat{\Lambda}}f$, $\nabla_{\hat{B}}f$, and $\nabla_{\hat{C}}f$ of $f(\hat{\Lambda}, \hat{B}, \hat{C})$ are given by
    \begin{align}
        &\nabla_{\hat{\Lambda}}f = 2\diag(Y_{\tau}^{*}X+\hat{Q}_{\tau}\hat{P} + \tau ( \mathcal{L}(\hat{A}\tau, S_{\tau})^{*})),\\
        &\nabla_{\hat{B}}f = 2(Y_{\tau}^{*}B+\hat{Q}_{\tau}\hat{B}),\quad  \nabla_{\hat{C}}f = 2(-CX_{\tau}+\hat{C}\hat{P}_{\tau}), 
    \end{align}
    where $\hat{P}_{\tau}$, $\hat{Q}_{\tau}$, $X_{\tau}$, and $Y_{\tau}$ are the solutions of \eqref{eq:f-lyap1}, \eqref{eq:f-lyap2}, \eqref{eq:f-sylve1}, and \eqref{eq:f-sylve2}, respectively, 
    and $\hat{P}$, $X$ are the solutions of
    \begin{align}
        \hat{A}\hat{P} + \hat{P}\hat{A}^{*} + \hat{B}\hat{B}^{*} = 0, \label{eq:i-lyap}\\
        AX + X\hat{A}^{*} + B\hat{B}^{*} = 0. \label{eq:i-sylve}
    \end{align}
    Here, $S_{\tau} := X^{*}\E^{A^{*}\tau}C^{*}\hat{C} - \hat{P}\E^{\hat{A}^{*}\tau}\hat{C}^{*}\hat{C}$, and $\mathcal{L}(\hat{A}, S_{\tau})$ denotes the Fr\'{e}chet derivative of the matrix exponential, defined as 
    $\mathcal{L}(\hat{A}, S_{\tau}) := \int_0^1 \E^{\hat{A}(1-s)} S_{\tau} \E^{\hat{A}s} \, ds$.
\end{theorem}
\begin{proof}
    Following~\cite{van2008h2, das2022h}, we derive the gradients using perturbation techniques. See Appendix~\ref{app:proof_of_theorem1}.
\end{proof}

\subsection{Gradients-based Algorithm for \eqref{eq:opt_modified}}\label{subsec:algorithm}

% アルゴリズムの説明
%--- Algorithm description (compact, minimal edits) --------------------------
% The optimization algorithm using the gradients derived in
% Section~\ref{subsec:gradients} is presented in
% Algorithm~\ref{alg:complex_h2_mor}. 
Algorithm~\ref{alg:complex_h2_mor} outlines the gradient-based optimization from Section~\ref{subsec:gradients}.
\tblack{Let $\phi_k := \paren{\hat{\Lambda}_k,\hat{B}_k,\hat{C}_k}$ and $\nabla f(\phi_k):=\paren{\nabla_{\hat{\Lambda}}f(\phi_{k}), \nabla_{\hat{B}}f(\phi_{k}), \nabla_{\hat{C}}f(\phi_{k})}$.
For each iteration $k=0,1,2,\dots$ update
$\phi_{k+1}=\phi_k-\alpha_k\nabla f(\phi_k)$,
where the step size $\alpha_k>0$ is chosen by backtracking until both the
Armijo condition and the stability constraint are satisfied.
Let $\mathcal{E}:=\{\hat A\in\mathbb{C}^{r\times r}\mid
\Re\lambda_i(\hat A)<0\ \forall i\}$; since $\mathcal{E}$ is open
\cite{orbandexivry2013nearest}, if
$\hat A_k=\operatorname{diag}(\hat{\Lambda}_k)\in\mathcal{E}$ then for
sufficiently small $\alpha_k$ we have
$\operatorname{diag}(\hat{\Lambda}_k-\alpha_k\nabla_{\hat{\Lambda}}f(\hat{\Lambda}_k))\in\mathcal{E}$.
The algorithm terminates when
$D_k:=\|\nabla_{\hat{\Lambda}}f(\phi_k)\|+
      \|\nabla_{\hat{B}}f(\phi_k)\|+
      \|\nabla_{\hat{C}}f(\phi_k)\|<\mathit{tol}$.}

% アルゴリズムの計算量
\begin{theorem}\label{thm:cost}
% \tblack{Assuming that the number of backtracking trials is $\mathcal{O}(\ell)$ and the maximum number of iterations is $K_{\max}$, 
\tblack{Assuming that the computational cost of the backtracking per iteration is $\ell$ and the maximum number of iterations is $K_{\max}$, the total cost of Algorithm~\ref{alg:complex_h2_mor} is}
\begin{align*}
    \tblack{\mathcal{O}\paren{\paren{Nr(r + m + p) + \ell}K_{\max}}.}
\end{align*}
\end{theorem}
\begin{proof}
    Since the matrix $A$ in \eqref{eq:c-SSM} is diagonal, the cost of the Lyapunov equations ~\eqref{eq:f-lyap1}, \eqref{eq:f-lyap2}, \eqref{eq:i-lyap} and Sylvester equations~\eqref{eq:f-sylve1}, \eqref{eq:f-sylve2}, \eqref{eq:i-sylve} are $\mathcal{O}(r^2)$ and $\mathcal{O}(Nr)$, respectively.
\end{proof}

% \begin{remark}\label{remark:diag_sylvester}
% Consider the Sylvester equation
% \begin{equation}
%     \mathcal{A} \mathcal{X} + \mathcal{X} \mathcal{B} = \mathcal{C}, \label{eq:diag_sylvester}
% \end{equation}
% where $\mathcal{A} \in \mathbb{C}^{N \times N}$ is stable and diagonal, and $\mathcal{B} \in \mathbb{C}^{r \times r}$ is diagonal.
% Let $\mathcal{X}_{i,j}$ denote the $(i,j)$-th entry of the solution matrix $\mathcal{X}$. Then, the explicit solution is given by
% \[
% \mathcal{X}_{i,j} = -\frac{\mathcal{C}_{i,j}}{\mathcal{A}_i + \mathcal{B}_j},
% \]
% where $\mathcal{A}_i$ and $\mathcal{B}_j$ are the $i$-th and $j$-th diagonal entries of $\mathcal{A}$ and $\mathcal{B}$, respectively. Therefore, the computational complexity of solving~\eqref{eq:diag_sylvester} is $\mathcal{O}(Nr)$.
% \end{remark}

% For the computational cost of each component, refer to Table~\ref{tab:cost}. 
% Since the matrix $A$ in the DSS model \eqref{eq:c-SSM} has a diagonal structure, both the Lyapunov equations ~\eqref{eq:f-lyap1}, \eqref{eq:f-lyap2}, \eqref{eq:i-lyap} and Sylvester equations~\eqref{eq:f-sylve1}, \eqref{eq:f-sylve2}, \eqref{eq:i-sylve} can be solved efficiently, as discussed in Remark~\ref{remark:diag_sylvester}. 
The main computational bottleneck of Algorithm \ref{alg:complex_h2_mor} is the matrix operations for evaluating the objective and its gradients.  
However, since the state dimension \(N\) in DDSSMs is typically \(\mathcal{O}(10^2)\), their cost is negligible compared to the overall training cost, as detailed in Remark~\ref{rem:cost}.

% % 計算量
% \begin{table}[!ht]
%   \centering
%   \caption{Computational complexity of each component in Algorithm~\ref{alg:complex_h2_mor}}
%   \label{tab:cost}
%   \scriptsize
%   \begin{tabular}{@{}>{\raggedright\arraybackslash}p{4.75cm}l@{}}
%     \toprule
%     \textbf{Component} & \textbf{Computational Complexity} \\
%     \midrule
%     $\mathrm{e}^{A\tau}BB^{*}\mathrm{e}^{A^{*}\tau}$ (matrix computations, computed once) 
%     & $\mathcal{O}(N^2 m)$ \\
    
%     $\mathrm{e}^{\hat{A}\tau}\hat{B}\hat{B}^{*}\mathrm{e}^{\hat{A}^*\tau}$ (matrix computations) 
%     & $\mathcal{O}(r^2 m)$ \\
    
%     Lyapunov equations~\eqref{eq:f-lyap1}, \eqref{eq:f-lyap2}, \eqref{eq:i-lyap} 
%     & $\mathcal{O}(r^2)$ \\
    
%     Sylvester equations~\eqref{eq:f-sylve1}, \eqref{eq:f-sylve2}, \eqref{eq:i-sylve} 
%     & \textbf{$\mathcal{O}(Nr)$} \\
    
%     Objective function~\eqref{eq:mod_obj} and gradients for Theorem~\ref{thm:gradients} and $S_{\tau}$ (matrix computations) 
%     & $\mathcal{O}(Nr^2 + mr^2 + Nmr + Npr)$ \\
    
%     $\mathcal{L}(\hat{A}_{\tau},S_{\tau})$ computation 
%     & $\mathcal{O}(r^2)$~\cite{al2009computing} \\
%     \bottomrule
%   \end{tabular}
% \end{table}

% アルゴリズム
\newpage
\noindent\mbox{}\vspace{1pt}
\begin{algorithm}[!htbp]
\caption{Complex diagonal finite-time $\mathcal{H}^{2}$-MOR}
\label{alg:complex_h2_mor}
\begin{algorithmic}[1]
\REQUIRE
  Initial ROM parameters $\hat{\Lambda}_{0}$, $\hat{B}_{0}$, $\hat{C}_{0}$, 
  tolerance $\mathit{tol}$, Armijo parameter $c_1$, initial step $\alpha_{\text{ini}}$, backtracking factor $\rho$, maximum number of iterations $K_{\max}$
\ENSURE
  ROM parameters $\hat{\Lambda}$, $\hat{B}$, $\hat{C}$ minimising \eqref{eq:mod_obj}
\FOR{$k = 0,1,\dots,K_{\max}$}  % <— 添字を 0-start の下付きに統一
    \STATE Solve \eqref{eq:f-lyap1}, \eqref{eq:f-lyap2}, \eqref{eq:i-lyap} for $\hat P_{\tau},\hat Q_{\tau},\hat P$
    \STATE Solve \eqref{eq:f-sylve1}, \eqref{eq:f-sylve2}, \eqref{eq:i-sylve} for $\hat Q_{\tau},Y_{\tau},\hat Q$
    \STATE Compute $f_k := f(\phi_k)$ and the gradients $\nabla f_k$
    \STATE \textbf{if} $D_k < \mathit{tol}$ \textbf{then break}

    \STATE $\alpha \leftarrow \alpha_{\text{ini}}$
    \WHILE{true}
        \STATE $\tilde{\phi} \leftarrow \phi_k - \alpha\,\nabla_{\phi}f_k$
        \STATE \textbf{if} $f(\tilde{\phi}) \le f_k - c_1\alpha D_k$ \textbf{and} $\tilde A$ is stable \textbf{then}\\
               \quad $\phi_{k+1} \leftarrow \tilde{\phi}$;\; \textbf{break}
        \STATE \textbf{else} $\alpha \leftarrow \rho\,\alpha$
    \ENDWHILE
\ENDFOR
\end{algorithmic}
\end{algorithm}
\vspace*{-5pt}

%%%%%%%%%%%%%
\section{Application to MOR-based Compression} \label{sec:experiments}

\begin{figure}[t]
    \vspace*{2mm} 
    \centering
    \includegraphics[width=8cm]{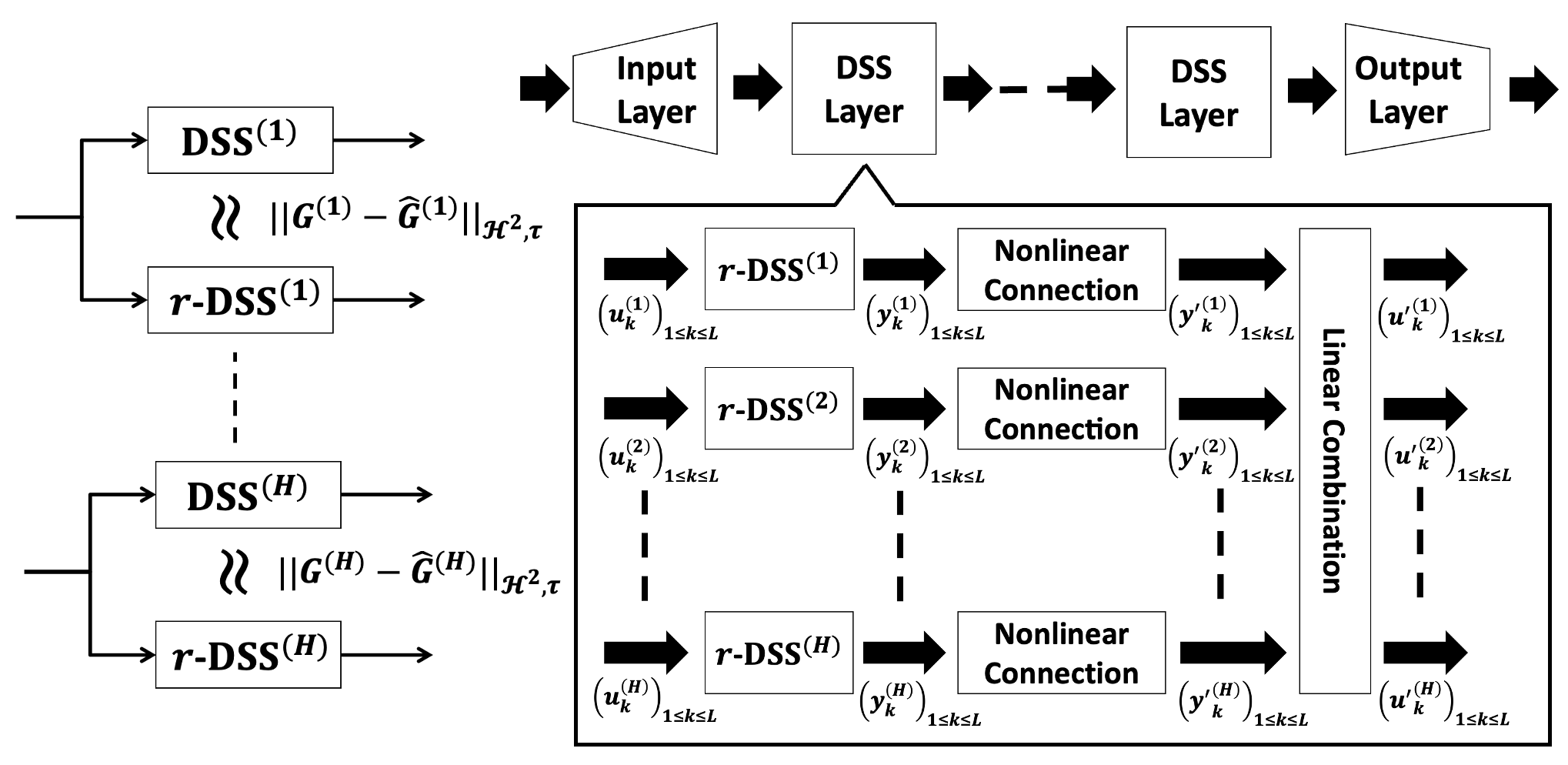}
    \caption{%
    % Construction of ROMs ($r$-DSS) via the proposed finite-time $\mathcal{H}^2$ MOR and a compressed model using the $r$-DSS. \quad
    \tblack{
    \textbf{Left:} Construction of $r$-dimensional ROMs ($r$-DSS) via the proposed finite-time $\mathcal{H}^2$ MOR. Starting from $N$-dimensional DSS models, we obtain $r$-DSS that is optimal in the finite-time $\mathcal{H}^2$ norm ($H$ is the number of DSS blocks per layer). \quad
    \textbf{Right:} The deep learning model architecture for re-training.
    We employ an $r$-DSS$_{\mathrm{EXP}}$ as $r$-DSS;
    the 1-D sequences \((u^{(i)}_{k})_{1\le k\le L}\) and
    \((y^{(i)}_{k})_{1\le k\le L}\) are mapped to
    \((u'^{(i)}_{k})_{1\le k\le L}\) and \((y'^{(i)}_{k})_{1\le k\le L}\) by nonlinear and linear combination.
    Note that the stability of $r$-DSS ensures that for any input $u_{k}$, the output $y_k$ does not diverge.
    See \cite{gupta2022diagonal,ezoe2024model} for details.
    }}
    \label{fig:ddssms_all}
\end{figure}

% 初めに
\tblack{We present the results of applying the proposed method from Section~\ref{sec:main} to the model compression framework of~\cite{ezoe2024model}, which is described in Section~\ref{subsec:compression}.
In this experiment, we employ the S4 architecture with DSS models \cite{gupta2022diagonal,ezoe2024model} as the DDSSM, based on the code available at \url{https://github.com/ag1988/dlr}.}
\tblack{See Figure~\ref{fig:ddssms_all} for the architecture with ROM obtained by the proposed method.}

% LRAベンチマークデータセットで提案手法を検証
We evaluate our compression method on the IMDb dataset~\cite{maas2011learning} from the LRA benchmark~\cite{tay2021long}, which targets long-context modeling.  
The task is binary sentiment classification with input sequence length \(L = 2048\).
\tblack{See Appendix~\ref{app:results_of_listops} for results using the ListOps dataset.}

As shown in Table~\ref{table:imdb}, the following models are compared by inference: the baseline model trained with Skew-HiPPO initialization~\cite{gu2020hippo, gupta2022diagonal} (HiPPO); compressed models based on the infinite-time BT~\cite{moore1981principal} (iBT), finite-time BT~\cite{gawronski1990model, kurschner2018balanced} (fBT), infinite-time \(\mathcal{H}^2\)-MOR (iH2), and the proposed finite-time \(\mathcal{H}^2\)-MOR (fH2).  
% For the finite-time MOR methods, performance is evaluated using three time horizons: \(\tau = L\Delta\), \(\tau = L\), and \(\tau = 10L\), where \(L\) is the input length and \(\Delta\) is the sample time. \tblack{Note that a discrete SSM~\eqref{eq:d-SSM} of length $L$ is transformed into a finite-time continuous SSM~\eqref{eq:c-SSM} with \(\tau = L\Delta\).}
% ここで，IMDbタスクでは学習，推論ともに入力系列長Lは同じである一方で，一般的には，Lと推論時の系列長Linfは必ずしも一致しない．本実験では，三つのパターンのtime-horizonで定義される有限時間SSMを考え，各々に有限時間MOR手法を適用する．
\tblack{Although the IMDb task fixes the input length at \(L\) for both training and inference, practical cases often have an inference length \(L_{\mathrm{inf}}\!\neq\!L\). To study how the horizon affects performance, we convert the discrete‐time SSM to a continuous‐time SSM under three horizons—\(\tau=L\Delta,\;L,\;10L\)—and apply finite-time MOR to each.}
% \tblack{Here, while the input length $L$ is the same for both training and inference in the IMDb task, in general, $L$ and the length $L_{\text{inf}}$ during inference do not necessarily coincide. In this experiment, we consider finite-time SSMs defined by three patterns of time-horizons  \(\tau = L\Delta\), \(\tau = L\), and \(\tau = 10L\) and apply the finite-time MOR to each of them.}

% % 実行環境
% The numerical experiments were conducted on a machine running Ubuntu~22.04.5~LTS, equipped with an Intel Xeon CPU (2.20~GHz, 6~cores / 12~threads), 168~GB of RAM, and a single NVIDIA A100-SXM4 GPU (80~GB, CUDA~12.8).

%%%%%%%%%%%%%
\subsection{DSS$_{\mathrm{EXP}}$ structure} 
\label{subsec:models}

According to~\cite{gu2020hippo,gupta2022diagonal}, high-performance models can be obtained by initializing DSS models~\eqref{eq:c-SSM} using Skew-HiPPO initialization~\cite{gu2020hippo, gupta2022diagonal}.
In particular, we use the following DSS$_{\mathrm{EXP}}$ models~\cite{gupta2022diagonal} as the DSS models:
\tblack{
\begin{subequations}\label{eq:dss-exp-model}
\begin{align}
    &A = \mathrm{diag}(\Lambda), \: B = \mathbf 1_{N}=\begin{bmatrix} 1 & \cdots & 1 \end{bmatrix}^{\top}, \: C = w^{\top},\label{eq:dss-exp}\\
    &\Lambda = -\exp(\Lambda_{re}) + \mathrm{i} \cdot \Lambda_{im}, \quad w = w_{re} + \mathrm{i} \cdot w_{im}, 
    \label{eq:lambda-w-exp}
\end{align}
\end{subequations}
}
where \(\Lambda_{re}, \Lambda_{im}, w_{re}, w_{im}\in \mathbb{R}^{N}\) are the parameters in training.
With the DSS$_{\mathrm{EXP}}$ structure, $\hat{A}$ is always stable due to the form in~\eqref{eq:lambda-w-exp}, and thus the stability constraint in Algorithm~\ref{alg:complex_h2_mor} is always satisfied.
Furthermore, since $B$ is constant, it is not updated during each iteration of the algorithm.
\tblack{Note that the proposed method in Section~\ref{sec:main} can be applied to SISO systems such as \eqref{eq:dss-exp} by setting $m=p=1$.}

% DSSexpの場合の勾配
\begin{prop}\label{prop:dss_exp_grad}
    Consider the DSS$_{\mathrm{EXP}}$~\eqref{eq:dss-exp-model} structure for \eqref{eq:c-SSM} and \eqref{eq:r-dss}. Then, the gradients for \eqref{eq:opt_modified} can be rewritten as:
    \begin{align*}
        &\nabla_{\hat{\Lambda}_{re}} f = - \nabla_{\Re(\hat{\Lambda})} f \odot \exp(\Re(\hat{\Lambda})),\quad \nabla_{\hat{\Lambda}_{im}} f = \nabla_{\Im(\hat{\Lambda})} f, \\
       &\nabla_{\hat{w}_{re}} f = \nabla_{\Re(\hat{C})} f, \quad\nabla_{\hat{w}_{im}} f = \nabla_{\Im(\hat{C})} f,
    \end{align*}
    where $\hat{\Lambda}_{re}$, $\hat{\Lambda}_{im}$, $\hat{w}_{re}$, and $\hat{w}_{im}$ are as defined in~\eqref{eq:lambda-w-exp}. 
\end{prop}
\begin{proof}
    \tblack{See Appendix~\ref{app:proof_of_prop2}.}
    % The proof is completed by applying the chain rule.
\end{proof}

%%%%%%%%%%%%%
\subsection{\tblack{Flow of $\mathcal{H}^2$ MOR-based Compression}} \label{subsec:compression}
Following~\cite{ezoe2024model}, the compressed model for the DDSSM is constructed through a three-stage process consisting of pre-training, solve the $\mathcal{H}^2$-MOR problem, and re-training.

% 事前学習の流れ
\subsubsection*{Pre-Training}
The initial parameters $(\Lambda, w)$ and $\Delta$ of the DSS$_{\mathrm{EXP}}$ models \eqref{eq:dss-exp-model} are determined by the Skew-HiPPO initialization~\cite{gupta2022diagonal}, after which the model is trained.

% モデル低次元化
\subsubsection*{Solve the $\mathcal{H}^2$-MOR problem}
We perform $\mathcal{H}^2$-MOR for the DSS$_{\mathrm{EXP}}$ models~\eqref{eq:dss-exp-model} obtained from the pre-training phase.
To execute Algorithm~\ref{alg:complex_h2_mor} using the gradients in Proposition~\ref{prop:dss_exp_grad}, initial ROM parameters are required.  
In this study, we use those obtained by the BT method as initial values.  
This initialization enables Algorithm~\ref{alg:complex_h2_mor} to construct ROMs with better finite-time \(\mathcal{H}^{2}\)-norm performance than those from the BT method.
Here, the sample time \(\Delta\) is fixed.

% 再学習
\subsubsection*{Re-Training}
The parameters \(\hat{\Lambda}_{re}, \hat{\Lambda}_{im}, \hat{w}_{re}, \hat{w}_{im} \in \mathbb{R}^{r}\) obtained by Algorithm~\ref{alg:complex_h2_mor} are used to initialize the DSS model in re-training, while all other parameters are initialized using those from the pre-training phase.

\begin{remark}\label{rem:cost}
    Let \(n_{\text{data}}\), \(n_{\text{epoch}}\), and \(B\) be the number of training samples, epochs, and batch size, respectively. 
    In training, each DSS model performs \(n_{\text{data}} n_{\text{epoch}} / B\) input-output operations, each costing \(\mathcal{O}(L \log L)\) using the FFT~\cite{cormen2022introduction}. 
    Thus, the total training cost per DSS model is \(\mathcal{O}(n_{\text{data}} n_{\text{epoch}} L \log L / B)\).
    By Theorem~\ref{thm:cost}, the computational bottleneck of Algorithm~\ref{alg:complex_h2_mor} is \(\mathcal{O}(N^2 K_{\max})\). When \(N = \mathcal{O}(10^2)\), this cost is negligible compared to training when \(n_{\text{data}}\) and $L$ are sufficiently large.
\end{remark}

%%%%%%%%%%%%%
\subsection{Pre-Training and $\mathcal{H}^{2}$ model order reduction} \label{subsec:pretrain}

\begin{table*}[t]
\centering
\small
\caption{Accuracy of models through various training methods
         (DSS$_\text{EXP}$, $H{=}128$, $\xi{=}4$).}
\label{table:imdb}

\begin{tabular}{
    c            % r
    S            % HiPPO
    *{2}{S}      % iBT
    *{2}{S}      % iH2
    *{2}{S}      % fH2(LΔ)
    *{2}{S}      % fH2(L)
    *{2}{S}      % fH2(10L)
}
\toprule
$r$ (or $N$) & \multicolumn{1}{c}{HiPPO}
    & \multicolumn{2}{c}{\textbf{iBT, fBT($L$), fBT($10L$)}}
    & \multicolumn{2}{c}{\textbf{iH2}}
    & \multicolumn{2}{c}{\textbf{fH2($L\!\Delta$)}}
    & \multicolumn{2}{c}{\textbf{fH2($L$)}}
    & \multicolumn{2}{c}{\textbf{fH2($10L$)}} \\
\cmidrule(lr){3-4}\cmidrule(lr){5-6}\cmidrule(lr){7-8}%
\cmidrule(lr){9-10}\cmidrule(lr){11-12}
 & & {bef.} & {aft.}
   & {bef.} & {aft.}
   & {bef.} & {aft.}
   & {bef.} & {aft.}
   & {bef.} & {aft.} \\
\midrule
32 & \bfseries 0.8471 & 0.6226 & 0.8378 & 0.6451 & 0.8406 & 0.5192 & 0.8336 & 0.6304 & 0.8400 & 0.6304 & 0.8400 \\
16 & 0.8362 & 0.5310 & 0.8393 & 0.5058 & 0.8408 & 0.5022 & 0.8324 & 0.5062 & 0.8386 & 0.5062 & 0.8416 \\
 8 & 0.8381 & 0.5354 & 0.8436 & 0.5030 & 0.8403 & 0.5040 & 0.8389 & 0.5038 & 0.8387 & 0.5038 & 0.8371 \\
 4 & 0.8389 & 0.5994 & 0.8380 & 0.5014 & 0.8366 & 0.5044 & 0.8400 & 0.5042 & 0.8393 & 0.5042 & 0.8436 \\
 2 & 0.8324 & 0.6120 & 0.8346 & 0.5028 & 0.8406 & 0.5074 & \bfseries 0.8451
                     & 0.5078 & 0.8424 & 0.5078 & 0.8376 \\
\bottomrule
\end{tabular}
\end{table*}
% The pre-training results for SSMs with \(N \in \{2, 4, 8, 16, 32, 64\}\) are shown in the “HiPPO” column of Table~\ref{table:imdb}, 
Pre-training is performed on SSMs with \(N=64\) using Skew-HiPPO initialization,
\tblack{where \(H = 128\) and \(\xi = 4\) denotes the number of intermediate layers; all other experimental settings follow~\cite{gupta2022diagonal}.}
% Comparing the models before and after pre-training (Table~\ref{table:imdb}, “before” and “after” columns) shows improved accuracy after pre-training, with \( N = 32 \) achieving the best performance.  
For reference, \cite{gupta2022diagonal} reported 84.6\% accuracy for DSS$_{\mathrm{EXP}}$ with \( N = 64 \), while our environment yielded 84.49\%.
Note that the results for SSMs with \(N \in \{2, 4, 8, 16, 32\}\) are shown in the “HiPPO” column of Table~\ref{table:imdb}.

% 初めに
After pre-training, \(\mathcal{H}^2\)-MOR is applied to 512 (\(= H \times \xi\)) DSS$_{\mathrm{EXP}}$ models and \( N = 64 \), using Algorithm~\ref{alg:complex_h2_mor} based on the gradients from Proposition~\ref{prop:dss_exp_grad}. 
\tblack{The initial ROMs are built with BT; when finite-time BT may not produce a stable ROM, a random stable model is constructed by randomly generating $\Lambda$ and $w$ in the DSS$_{\mathrm{EXP}}$ models~\eqref{eq:dss-exp-model}.}
% \tblack{If the initial ROM by finite-time BT is unstable, we can use use alternative MOR methods—e.g., random initialization by randomly generating $\Lambda$ and $w$ in the DSS$_{\mathrm{EXP}}$ models~\eqref{eq:dss-exp-model} or Krylov methods~\cite{antoulas2005approximation}.}
The optimization parameters are set as: \(\mathit{tol} = 10^{-3}\), \(c_1 = 10^{-4}\), \(\rho = 0.5\), \(K_{\max} = 100\), and \(\alpha_{\text{ini}} = 1\).

\begin{figure}[htbp]
    \centering
    \subfloat[Infinite-Time\label{fig:inf}]{%
    \includegraphics[width=0.49\linewidth, height=2.9cm]{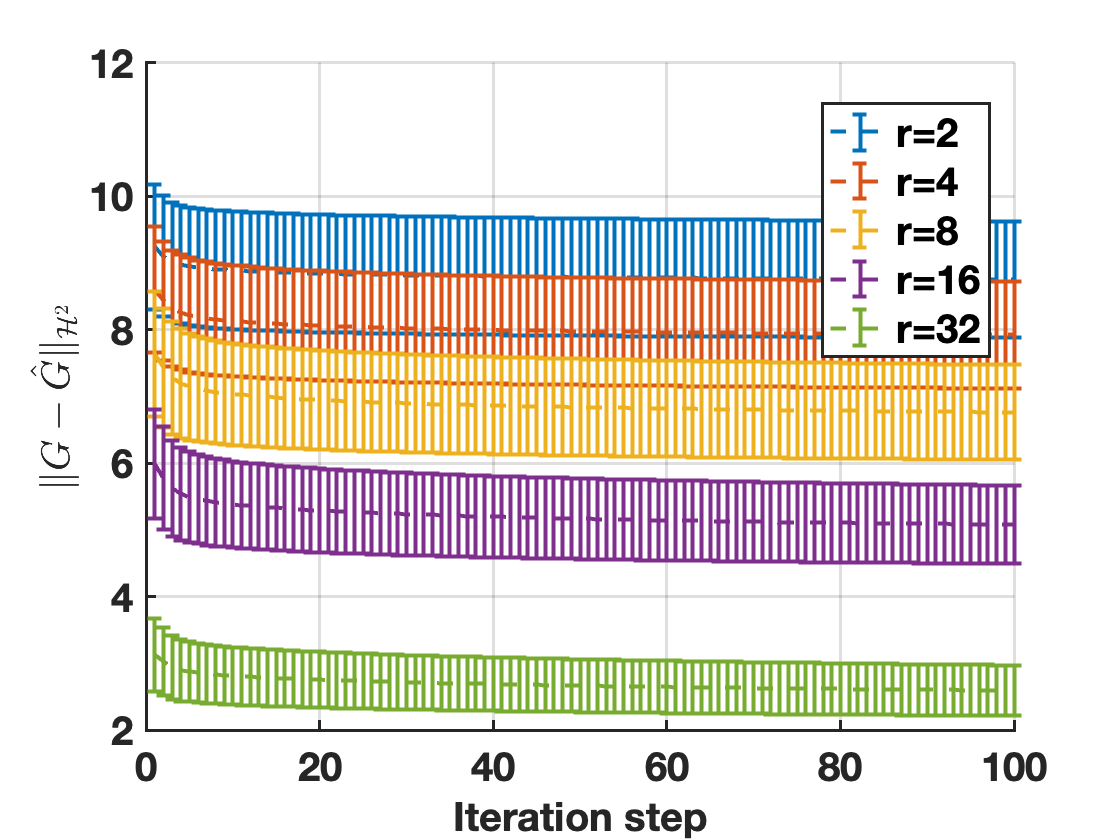}%
    }\hfill
    \subfloat[$\tau=10L$\label{fig:10L}]{%
    \includegraphics[width=0.49\linewidth, height=2.9cm]{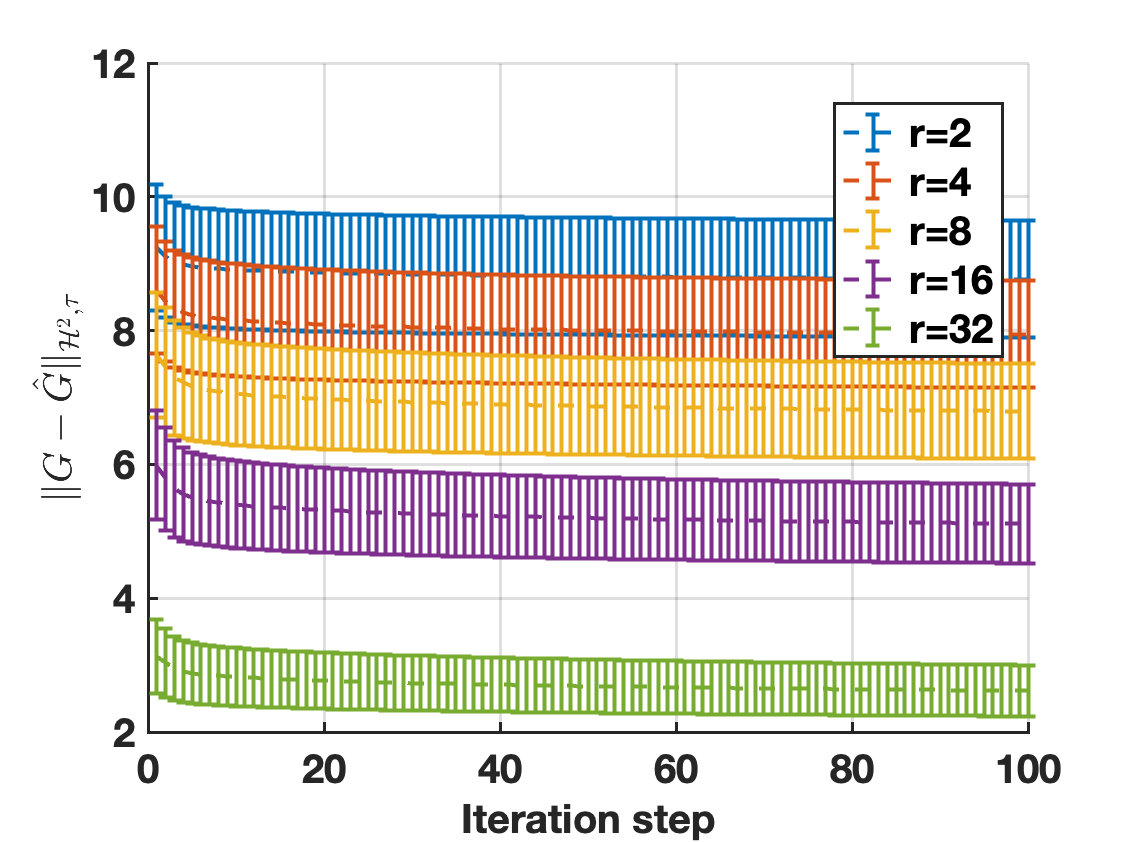}%
    }\\
    \subfloat[$\tau=L$\label{fig:L}]{%
    \includegraphics[width=0.49\linewidth, height=2.9cm]{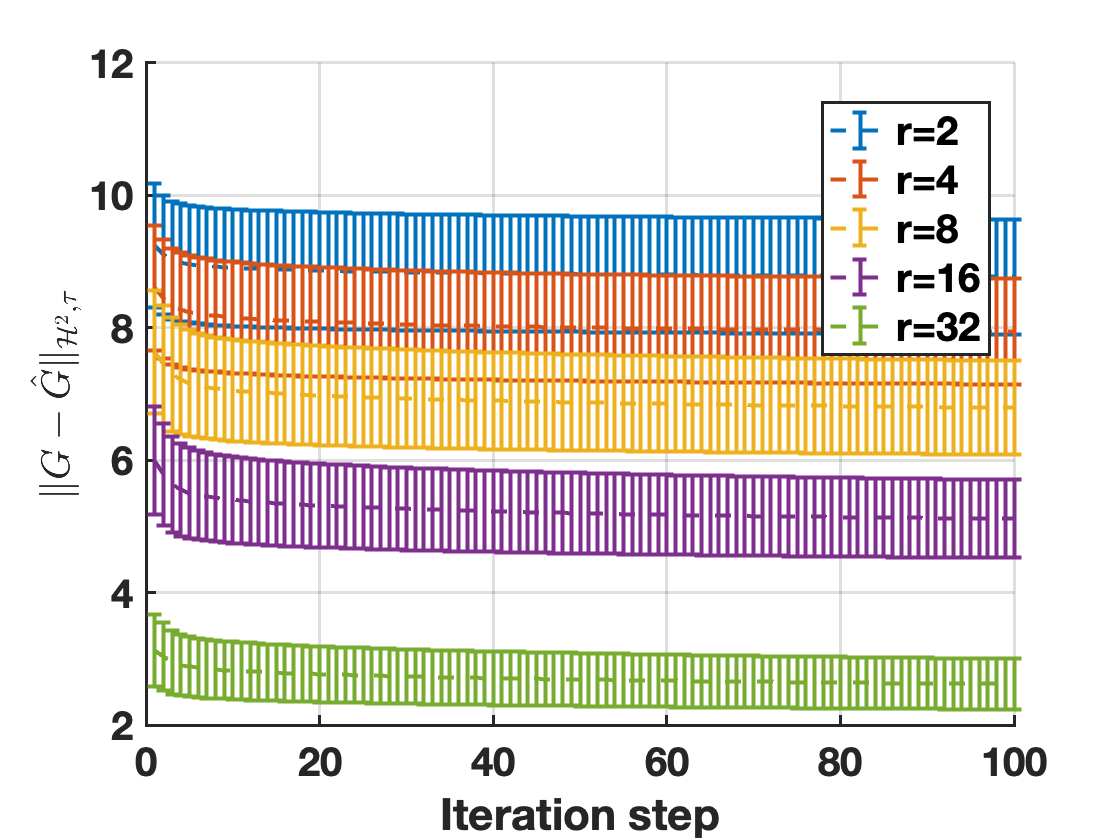}%
    }\hfill
    \subfloat[$\tau=L\Delta$\label{fig:Ldt}]{%
    \includegraphics[width=0.49\linewidth, height=2.9cm]{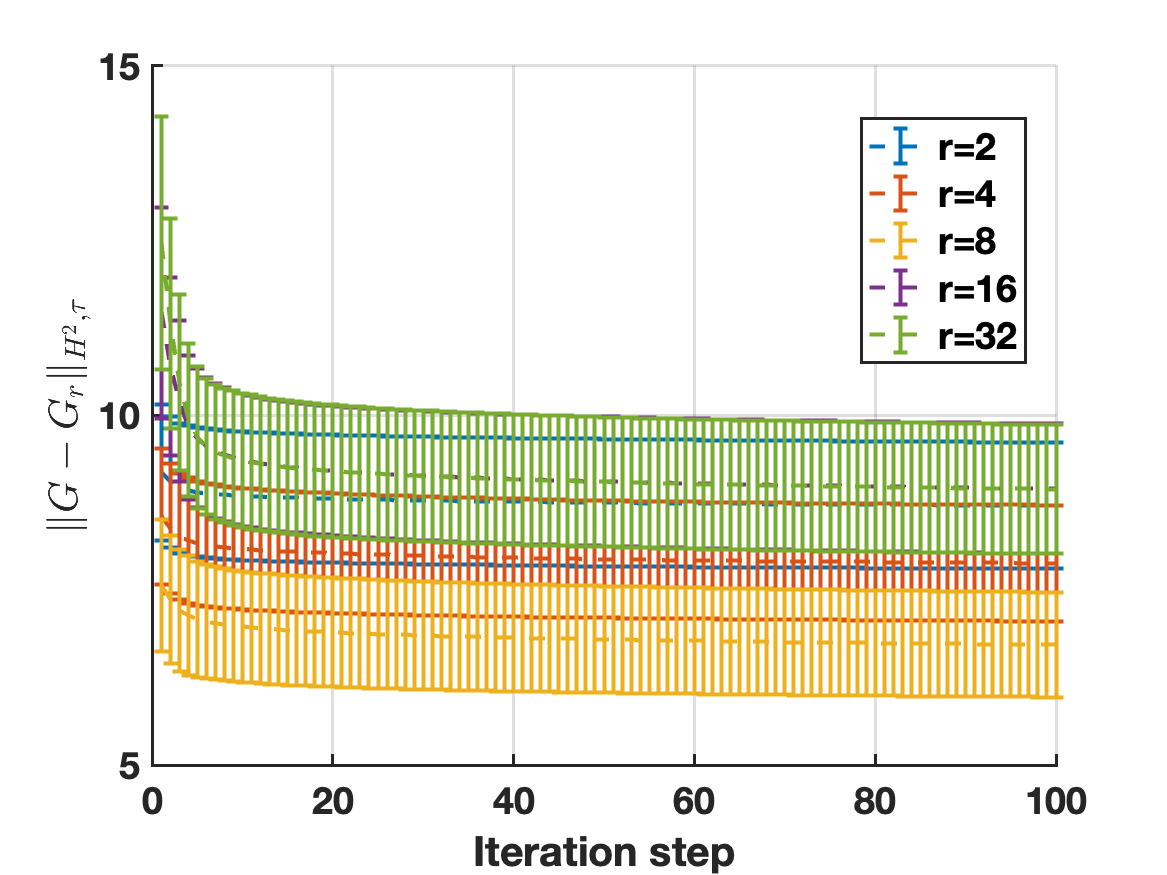}%
    }
    \caption{Convergence behavior of Algorithm~\ref{alg:complex_h2_mor}.
    (a) shows the result for the infinite-time \(\mathcal{H}^{2}\)-MOR problem with an infinite-time BT initialization.
    (b)–(d) show the results for the finite-time problem~\eqref{eq:opt_original} with \(\tau = 10L\), \(L\), and \(L\Delta\), respectively, using the corresponding finite-time BT initializations;
    \tblack{in (d), however, the cases \(r=16\) and \(r=32\) are initialized with random stable systems because of the instability of ROMs by finite-time BT.}}
    \label{fig:convergence}
\end{figure}

% 結果の説明
Figure~\ref{fig:convergence} shows the mean \(\pm\) standard deviation of the objective function values over the 512 DSS$_{\mathrm{EXP}}$ models.  
\tblack{The results demonstrate that, for all methods and \(r\), the proposed algorithm produces ROMs with \(\text{DSS}_{\text{EXP}}\) structure that achieve lower finite-time \(\mathcal{H}^2\) loss than the initial ROMs.}

% 収束の挙動を表す図
\vspace*{-5pt}

%%%%%%%%%%%%%
\subsection{Evaluation and Analysis of $\mathcal{H}^{2}$ MOR-Based Compression} \label{subsec:h2mor-compression}

% 図・結果の説明
Table~\ref{table:imdb} shows the model accuracy for several methods. 
\tblack{Here, the ROMs by fBT for $\tau=L$ and $10L$ and iBT almost coincide, as $\tau$ is sufficiently large. Thus, the accuracy of the compression models by iBT, fBT($L$) and fBT($10L$) coincided.}
\tblack{Note also that fH2($L\Delta$) for $r=16,32$ uses random stable systems as the initial point.}
In the case of fBT($L\Delta$), the ROMs became unstable, and when such unstable models were used as the initial models, no performance improvement was observed.

From Table~\ref{table:imdb}, it can be observed that using ROMs that better approximate the original system in the $\mathcal{H}^{2}$ norm as the initialization for re-training a deep model leads to relatively accurate models.  
\tblack{See Appendix~\ref{app:detailed_experiments_imdb} for detailed experiments on the effectiveness of the finite-time $\mathcal{H}^{2}$-MOR for model compression.}
In particular, among the compressed models with $r \le 32$, fH2($L\Delta$) for $r = 2$ achieved the best performance (84.51\%), outperforming the deep learning model with $N = 64$ (84.49\%) while reducing the number of state-space parameters to $\tfrac{1}{32}$ of the original.

% t検定の結果
\tblack{We then assessed whether fH2($L\Delta$) surpasses iBT by conducting two-sided \(t\)-tests on the same 10 random seeds. Across the 10 seeds, the test gave \(t=3.30,\,p=0.0093<0.01\), showing that fH2($L\Delta$) significantly outperforms iBT at \(r=2\); it also surpasses both the random stable models and fH2($L\Delta$) initialized from those random models.}
\tblack{See Appendix~\ref{app:t_tests} for details on the \(t\)-tests.}

% 考察
Compared with the infinite-time BT compression~\cite{ezoe2024model}, the superior accuracy of the $\mathcal{H}^{2}$-MOR compression stems from starting re-training with a higher-quality model. Two hypotheses support this: (i) the frequency responses of the pre-trained DSS models were well approximated, giving strong initial models; and (ii) these SSMs were handled as finite-time, not infinite-time, systems during MOR. Consequently, the proposed method is a reliable alternative when the infinite-time BT approach is not performing.

\section{Conclusion} \label{sec:conclusion}
% まとめ
In this study, we propose an $\mathcal{H}^2$-MOR for SSMs with the specific properties observed in 
% deep learning architectures that incorporate diagonal SSMs. 
DDSSMs.
Using the LRA benchmark, we show that our model compression approach outperforms the BT-based compression method~\cite{ezoe2024model}. Furthermore, our results demonstrate that it is possible to reduce the number of parameters in the SSMs to $1/32$ of its original size while preserving the performance of large-scale deep models.

% % 今後の課題
% \tblack{Algorithm~\ref{alg:complex_h2_mor} can also be applied to \(\mathrm{DSS}_{\mathrm{SOFTMAX}}\) systems~\cite{gupta2022diagonal} and to the MIMO systems that appear in S5~\cite{smith2023simplified}. Future work will explore its application and effectiveness on these settings.}
\tblack{In future work, we consider the application of the proposed $\mathcal{H}^2$-MOR method to tasks where it has more intrinsic advantages. Specifically, we consider its application to multi-step prediction tasks using time-series data obtained from physical systems described in continuous time.}

\section*{Acknowledgment}
\tblack{This work was supported by JSPS KAKENHI under Grant Numbers 23K28369 and 25KJ0986.}

\appendices
% \appendix
%%%%%%%%%%%%%
\section*{APPENDIX}
\makeatletter
\renewcommand{\appendixname}{}
\renewcommand{\thesectiondis}[2]{\Alph{section}.}
\makeatother

\section{Proof of Proposition~\ref{prop:modified}}\label{app:proof_of_prop1}
\begin{proof}
    Following~\cite[Proposition~2.2]{goyal2019time} and using the error system between \eqref{eq:c-SSM} and \eqref{eq:r-dss}, we obtain
    \begin{align*}
    \|G - \hat{G}\|_{\mathcal{H}^{2},\tau}^{2} &= \mathrm{tr} \left(
    \begin{bmatrix}
    B^{*} & \hat{B}^{*}
    \end{bmatrix}
    \begin{bmatrix}
    Q_{\tau} & Y \\
    Y^* & \hat{Q}_{\tau}
    \end{bmatrix}
    \begin{bmatrix}
    B \\ \hat{B}
    \end{bmatrix}
    \right) \\
    &= \tr (B^{*}Q_{\tau}B + \hat{B}^{*}\hat{Q}_{\tau}\hat{B} + 2\Re(\hat{B}^{*}Y_{\tau}^{*}B)).
    \end{align*}
    Similarly, 
    \begin{align*}
    \|G - \hat{G}\|_{\mathcal{H}^{2},\tau}^{2} &= \tr \left(
    \begin{bmatrix}
    C & -\hat{C}
    \end{bmatrix}
    \begin{bmatrix}
    P_{\tau} & X \\
    X^* & \hat{P}_{\tau}
    \end{bmatrix}
    \begin{bmatrix}
    C^{*} \\ -\hat{C}^{*}
    \end{bmatrix}
    \right) \\
    &= \tr (CP_{\tau}C^{*} + \hat{C}\hat{P}_{\tau}\hat{C}^{*} - 2\Re(\hat{C}X_{\tau}^{*}C^{*})).
    \end{align*}
\end{proof}

\section{Evaluation metrics for the BT and the proposed method}\label{app:evaluation_metrics}
% BTの評価指標
The ROM $\hat{G}_{\text{BT}}$ obtained by BT is good in the sense of the $\mathcal{H}^\infty$ norm and the well-known inequality holds: 
\begin{align}\label{eq:BT_inequality}
    \sigma_{r+1} \leq \| G - \hat{G}_{\text{BT}} \|_{\mathcal{H}^\infty} \leq 2 (\sigma_{r+1} + \cdots + \sigma_N),
\end{align}
where $\sigma_{r}$ is the $r$-th Hankel singular value and $\sigma_{r+1} > \cdots > \sigma_N$ \cite{antoulas2005approximation}.
The BT does not guarantee optimality in the sense of the $\mathcal{H}^\infty$ norm or the $\mathcal{H}^2$ norm.

% Proposed methodの評価指標  
Our approach seeks a ROM that is optimal in the finite-time $\mathcal{H}^2$ norm, motivated by the relation~\eqref{eq:relation-g-y}.
Unlike the BT, the proposed method enables the construction of ROMs with guaranteed $\mathcal{H}^2$-optimality~\cite{goyal2019time, sinani2019h2, das2022h}. 
Thus, it is difficult to discuss the input-output relationship in the same way as the BT, since the BT and the proposed method have different evaluation metrics to look at. 

\section{Proof of Theorem \ref{thm:gradients}}\label{app:proof_of_theorem1}
\begin{proof}
    From $\tr(\Re(\cdot))=\Re(\tr(\cdot))$ and \eqref{eq:mod_obj}, the first-order perturbation $\Delta^{\Re(\hat{\Lambda})}_{f}$ corresponding to $\Delta_{\Re(\hat{\Lambda})}$ is given by
    $\Delta^{\Re(\hat{\Lambda})}_{f} = 2\Re (\tr (\hat{B}B^{*}\Delta_{Y_{\tau}})) + \tr (\hat{B}\hat{B}^{*}\Delta_{\hat{Q_{\tau}}}),$
    where $\Delta_{Y_{\tau}}$ and $\Delta_{\hat{Q_{\tau}}}$ are the perturbations in $Y_{\tau}$ and $\hat{Q_{\tau}}$.
    Here, from \eqref{eq:f-sylve2}, 
    \begin{align}
        A^{*}\Delta_{Y_{\tau}}+\Delta_{Y_{\tau}}\hat{A} + Y_{\tau}\Delta_{\Re(\hat{A})}+\E^{A^{*}\tau}C^{*}\hat{C}\Delta_{\E^{\hat{A}\tau}} = 0,\label{eq:pert-f-sylve2}
    \end{align}
    where $\Delta_{\E^{\hat{A}\tau}} = \mathcal{L}\paren{\E^{\hat{A}\tau},\tau\Delta_{\Re(\hat{A})}}$.
    Similarly, from \eqref{eq:f-lyap2}, 
    \begin{align}
        \hat{A}^{*}\Delta_{\hat{Q}_{\tau}}+\Delta_{\Re(\hat{A})}^{*}\hat{Q}_{\tau} + \hat{Q}_{\tau}\Delta_{\Re(\hat{A})} + \Delta_{\hat{Q}_{\tau}}\hat{A} \\
        - \Delta_{\E^{\hat{A}\tau}}^{*}\hat{C}^{*}\hat{C}\E^{\hat{A}\tau} - \E^{\hat{A}^{*}\tau}\hat{C}^{*}\hat{C}\Delta_{\E^{\hat{A}\tau}} = 0. \label{eq:pert-f-lyap2}
    \end{align}
    Furthermore, from \cite[Lemma 3.2]{van2008h2}, \eqref{eq:f-lyap1}, \eqref{eq:f-lyap2}, \eqref{eq:pert-f-sylve2}, and \eqref{eq:pert-f-lyap2},
    \begin{align*}
        &\tr (Y_{\tau}\Delta_{\Re(\hat{A})}X^{*} + \E^{\hat{A}^{*}\tau}\hat{C}^{*}\hat{C}\Delta_{\E^{\hat{A}\tau}}) = \tr(\hat{B}B^{*}\Delta_{Y_{\tau}}).\\
        &\tr\paren{\paren{2\Re\paren{\hat{Q}_{\tau}\Delta_{\Re(\hat{A})}} - 2\Re \paren{ \E^{\hat{A}^{*}\tau}\hat{C}^{*}\hat{C}\Delta_{\E^{\hat{A}\tau}}}}\hat{P}} \\
        &= \tr (\hat{B}\hat{B}^{*}\Delta_{\hat{Q}_{\tau}}).
    \end{align*}
    Note that $\hat{Q}_{\tau}=\hat{Q}_{\tau}^{*}$ holds since $\hat{A}$ is stable.
    Thus, 
    \begin{align*}
        \Delta_{f}^{\Re(\hat{A})} &= 2\Re\paren{\tr \paren{\paren{X^{*}Y_{\tau} + \hat{P}\hat{Q}_{\tau}} \Delta_{\Re(\hat{A})} +S_{\tau}\Delta_{\E^{\hat{A}\tau}}}}.
    \end{align*}
    Furthermore,
    \begin{align*}
        \tr\paren{S_{\tau}\Delta_{\E^{\hat{A}\tau}}}
        &= \tr\paren{S_{\tau}\cdot\mathcal{L}\paren{\E^{\hat{A}\tau},\tau\Delta_{\hat{A}}}}  \\
        % &= \tr\paren{S_{\tau}\int_{0}^{1}\exp(\E^{\hat{A}\tau}(1-s))\cdot\tau\Delta_{\Re(\hat{A})} \cdot \exp(\E^{\hat{A}\tau}s) ds}  \\
        &= \tau \cdot \tr \paren{\mathcal{L}\paren{\hat{A}\tau,S_{\tau}}\cdot \Delta_{\Re(\hat{A})}}.
    \end{align*}
    Therefore, the following equation holds: 
    \begin{align*}
        \Delta_{f}^{\Re(\hat{A})} = \left\langle 2\Re \paren{Y_{\tau}^{*}X+\hat{Q}_{\tau}\hat{P} + \tau\cdot \mathcal{L}\paren{\hat{A}\tau,S_{\tau}}^{*}}
        , \Delta_{\Re(\hat{A})}
        \right\rangle.
    \end{align*}
    Since $\Delta_{\Re(\hat{A})}$ is diagonal, 
    \begin{align*}
        \nabla_{\Re(\hat{\Lambda})}f = 2\diag(\Re(Y_{\tau}^{*}X+\hat{Q}_{\tau}\hat{P} + \tau ( \mathcal{L}(\hat{A}\tau, S_{\tau})^{*})).
    \end{align*}
    Similarly, $\nabla_{\Im(\hat{\Lambda})}f$, $\nabla_{\Re(\hat{B})}f$, $\nabla_{\Im(\hat{B})}f$, $\nabla_{\Re(\hat{C})}f$, and $\nabla_{\Re(\hat{C})}f$ can be derived in the same manner.
\end{proof}

\section{Experiments on the ListOps dataset}\label{app:results_of_listops}
% ListOpsの説明
We report results on the ListOps dataset from the LRA benchmark~\cite{tay2021long}. ListOps is a synthetic list-operation task that evaluates a model's ability to handle hierarchical structure and long-range dependencies, with particular emphasis on correctly processing deeply nested lists.

% 事前学習の結果
In the experiment, the dimension of the pre-training SSM was set to $N=64$, and the other pre-training hyper-parameters were matched to those used in~\cite{gupta2022diagonal}. As a result, the accuracy after pre-training was 60.50\% in our environment.
For reference, \cite{gupta2022diagonal} reported 59.7\% accuracy for DSS$_\text{EXP}$ with $N=64$ and 60.6\% accuracy for DSS$_\text{SOFTMAX}$ with $N=64$.

% 再学習の結果
Table~\ref{table:listops} compares our method (fH2) with BT. 
Note that all $\mathcal{H}^2$‑based MOR are initialized with the output of BT.
As Table~\ref{table:listops} shows, the proposed method again surpasses the BT-based compression. In particular, for \(r = 4\) the fH2 configuration not only exceeds the pre-training accuracy (60.50\%), but also reduces the number of SSM parameters to $1/16$ of the original.

\begin{table*}[htbp]
  \centering
  \small
  \caption{ListOps accuracy after training
           (DSS$_\text{EXP}$, $H{=}128$, $\xi{=}6$).}
  \label{table:listops}

  \begin{tabular}{
      c            % r
      *{1}{S}      % iBT
      *{1}{S}      % fBT(LΔ)
      *{1}{S}      % fBT(L)
      *{1}{S}      % fBT(10L)
      *{1}{S}      % iH2
      *{1}{S}      % fH2(LΔ)
      *{1}{S}      % fH2(L)
      *{1}{S}      % fH2(10L)
  }
  \toprule
    $r$
    & \multicolumn{1}{c}{\textbf{iBT}}
    & \multicolumn{1}{c}{\textbf{fBT($L\!\Delta$)}}
    & \multicolumn{1}{c}{\textbf{fBT($L$)}}
    & \multicolumn{1}{c}{\textbf{fBT($10L$)}}
    & \multicolumn{1}{c}{\textbf{iH2}}
    & \multicolumn{1}{c}{\textbf{fH2($L\!\Delta$)}}
    & \multicolumn{1}{c}{\textbf{fH2($L$)}}
    & \multicolumn{1}{c}{\textbf{fH2($10L$)}} \\
  \midrule
    32 & 0.6085 & 0.1780 & 0.6085 & 0.6085 & 0.6035 & {-}    & 0.6045 & 0.6075 \\
    16 & 0.5990 & 0.1780 & 0.6045 & 0.5990 & 0.6000 & {-}    & 0.6080 & 0.6025 \\
     8 & 0.6000 & 0.1780 & 0.6000 & 0.6000 & 0.6030 & 0.6035 & 0.6005 & 0.6005 \\
     4 & 0.5975 & 0.1780 & 0.5975 & 0.5975 & 0.5975 & 0.6020 & \textbf{0.6090} & 0.5900 \\
     2 & 0.6045 & 0.1780 & 0.6045 & 0.6045 & 0.5995 & 0.6055 & 0.5900 & 0.6050 \\
  \bottomrule
  \end{tabular}
\end{table*}

\section{Proof of Proposition~\ref{prop:dss_exp_grad}}\label{app:proof_of_prop2}
\begin{proof}
    By the chain rule, 
    \begin{align*}
        \frac{\partial f}{\partial \hat{\Lambda}_{re,i}} 
        &= \sum_{j=1}^{r}\frac{\partial f}{\partial \Re(\hat{\Lambda})_{j}}\cdot \frac{\partial \Re(\hat{\Lambda})_{j}}{\partial \hat{\Lambda}_{re,i}} \\
        &= (\nabla_{\Re(\hat{\Lambda})}f)_{j}\cdot (-\exp (\Re(\hat{\Lambda})_{j})).
    \end{align*}
    Thus, $\nabla_{\hat{\Lambda}_{re}} f = - \nabla_{\Re(\hat{\Lambda})} f \odot \exp(\Re(\hat{\Lambda}))$.
\end{proof}

\section{Detailed experiments on the IMDb dataset}\label{app:detailed_experiments_imdb}
To evaluate the robustness and general applicability of our algorithm, we investigated several alternative initialization schemes in addition to the BT scheme for $r = 2, 4, \text{and}\ 8$. Specifically, we considered  
\begin{itemize}
  \item random stable ROMs (rand--ROM);  
  \item ROMs obtained by infinite-time BT (iBT--ROM);  
  \item ROMs obtained by running the Algorithm~\ref{alg:complex_h2_mor} from rand--ROM initial points (fH2--rand--ROM);  
  \item ROMs obtained by running the Algorithm~\ref{alg:complex_h2_mor} from fBT--ROM initial points (fH2--fBT--ROM), where fH2--fBT--ROM is the finite-time BT method.  
\end{itemize}

\begin{figure}[htbp]
  \centering
  \includegraphics[width=0.7\linewidth]{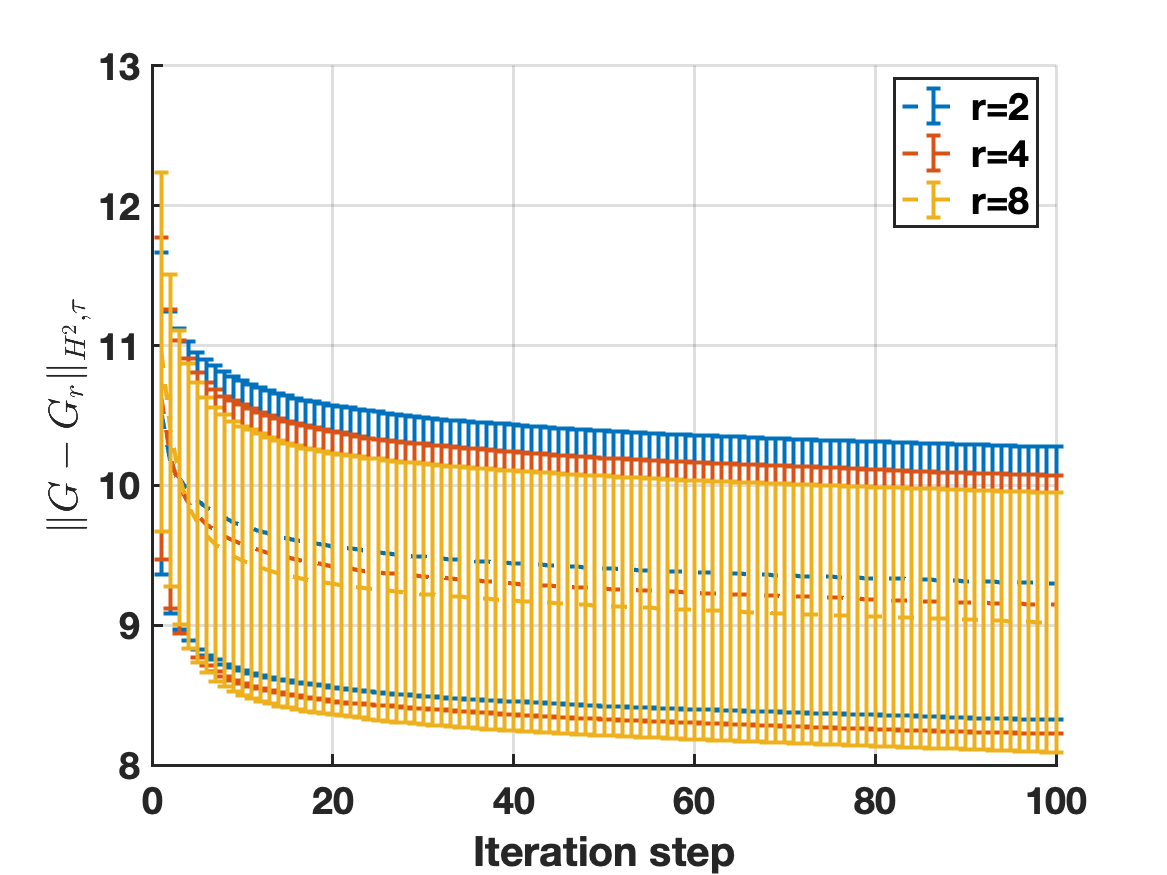}
  \caption{Convergence behavior of Algorithm~\ref{alg:complex_h2_mor} for fH2--rand--ROM.}
  \label{fig:H2_objective_evolution}
\end{figure}

For the rand--ROM, we generated the DSS$_{\text{EXP}}$ parameters 
$\Lambda_{re},\ \Lambda_{im},\ w_{re},\ w_{im}$ using \verb|randn| in \textsc{Matlab}. Note that the resulting ROMs are stable.  
For the fH2--rand--ROM and fH2--fBT--ROM, we set the time horizon to $\tau = L\Delta$.  
The iBT--ROM and fH2--fBT--ROM correspond to iBT and fH2($\tau = L\Delta$), respectively, as described in Section~\ref{sec:experiments}.

\begin{itemize}
  \item \textbf{Convergence behavior} ---  
  Figure~\ref{fig:H2_objective_evolution} plots the mean $\pm$ standard deviation of the objective function values over the $512$ DSS$_{\text{EXP}}$ models for the fH2--rand--ROM.  
  Relative to the rand--ROM baseline, our algorithm quickly yields a ROM with a significantly lower finite-time $\mathcal{H}^2$ loss.
  See Figure~\ref{fig:convergence} in Section~\ref{sec:experiments} for the objective trajectory for the fH2--fBT--ROM.
  \item \textbf{Performance of model compression} ---  
  Table~\ref{table:imdb_all_seeds} reports the accuracy of the test on the IMDb task. 
  Keeping every parameter identical except for the $r$-dimensional ROM and the random seed used at re-training time, we observe 
  \begin{itemize}
      \item fH2--rand--ROM consistently outperforms rand--ROM;  
      \item fH2--fBT--ROM attains the best overall accuracy, confirming the benefit of BT initialization for our optimization scheme.  
      % Furthermore, as described in \textbf{Reviewer 21--Comment~18}, a t-test showed that fH2--fBT--ROM was the best with $p<0.01$.
  \end{itemize}
\end{itemize}

\begin{table}[htbp]
  \caption{Accuracy on the IMDb dataset for five random seeds (0–4) and their mean (DSS$_{\text{EXP}}$, $H{=}128$, $\xi{=}4$). The model was pre‑trained with a state‑space dimension of $N{=}64$, yielding a baseline accuracy of $0.8449$.}
  \label{table:imdb_all_seeds}
  \centering
  \setlength{\tabcolsep}{4pt}       % compact columns
  \begin{tabular}{c c *{5}{c} c}
    \toprule
    $r$ & Method
        & Seed 0 & Seed 1 & Seed 2 & Seed 3 & Seed 4
        & Mean \\
    \midrule
    %---------------- N = 8 ----------------------------------------------------
    \multirow{4}{*}{8}
      & \textbf{rand--ROM}        & 0.8178 & 0.8250 & 0.8205 & 0.8207 & 0.8252 & 0.8218 \\
      & \textbf{iBT--ROM}         & 0.8390 & 0.8379 & 0.8381 & 0.8380 & 0.8374 & 0.8381 \\
      & \textbf{fH2--rand--ROM}   & 0.8372 & 0.8398 & 0.8347 & 0.8356 & 0.8430 & 0.8381 \\
      & \textbf{fH2--fBT--ROM}    & 0.8408 & 0.8427 & 0.8408 & 0.8432 & 0.8430 & \textbf{0.8421} \\
    \midrule
    %---------------- N = 4 ----------------------------------------------------
    \multirow{4}{*}{4}
      & \textbf{rand--ROM}        & 0.8026 & 0.8015 & 0.8057 & 0.8137 & 0.8090 & 0.8065 \\
      & \textbf{iBT--ROM}         & 0.8328 & 0.8390 & 0.8337 & 0.8445 & 0.8408 & 0.8382 \\
      & \textbf{fH2--rand--ROM}   & 0.8344 & 0.8386 & 0.8340 & 0.8405 & 0.8385 & 0.8372 \\
      & \textbf{fH2--fBT--ROM}    & 0.8416 & 0.8372 & 0.8372 & 0.8445 & 0.8330 & \textbf{0.8387} \\
    \midrule
    %---------------- N = 2 ----------------------------------------------------
    \multirow{4}{*}{2}
      & \textbf{rand--ROM}        & 0.8014 & 0.8111 & 0.8047 & 0.8028 & 0.8101 & 0.8060 \\
      & \textbf{iBT--ROM}         & 0.8343 & 0.8354 & 0.8427 & 0.8338 & 0.8356 & 0.8364 \\
      & \textbf{fH2--rand--ROM}   & 0.8371 & 0.8409 & 0.8435 & 0.8358 & 0.8356 & 0.8386 \\
      & \textbf{fH2--fBT--ROM}    & 0.8462 & 0.8360 & 0.8442 & 0.8419 & 0.8428 & \textbf{0.8422} \\
    \bottomrule
  \end{tabular}
\end{table}

In summary, BT provides a stronger initial point than purely random initialization for the Algorithm~\ref{alg:complex_h2_mor}.  Investigating other classical MOR initialisers, such as Krylov-based methods~\cite{antoulas2005approximation}, will be an interesting direction for future work.

\section{Comparison of the BT-based compression and the proposed method by the \(t\)-tests}\label{app:t_tests}
In this section, we compare iBT--ROM and fH2--fBT--ROM for $r=2$ using the \(t\)-tests.
First, we estimate the required sample size \(n_{\text{sample}}\) from the pilot experiment reported in Table~\ref{table:imdb_all_seeds} using power analysis. 
In Table~\ref{table:imdb_all_seeds}, five paired results were obtained, and the differences \(d_i\) between iBT--ROM and fH2--fBT--ROM for $r=2$ yielded an average \(\bar d = 0.00586\) and a standard deviation \(s_d = 0.00474\). 
Hence the paired-sample effect size is \(d_z = \bar d / s_d \approx 1.24\). Using a two-sided significance level of \(\alpha = 0.05\) and a statistical power of \(1 - \beta = 0.80\), with the standard normal quantiles \(z_{1-\alpha/2} = 1.96\) and \(z_{1-\beta} = 0.84\), the required number of pairs is
\[
n_{\text{sample}} = \left( \frac{z_{1-\alpha/2} + z_{1-\beta}}{d_z} \right)^2
    = \left( \frac{1.96 + 0.84}{1.24} \right)^2 \approx 5.2 .
\]
Taking the degrees of freedom into account, we need to collect at least 10 pairs (i.e., to run each method with 10 different seeds) so that the comparison between iBT--ROM and fH2--fBT--ROM for $r=2$ will achieve 80\% power.

We then present the results of a \(t\)-test 
$n_{\text{sample}}=10$ different random seeds.
Specifically, we present the results for “rand--ROM vs. fH2--fBT--ROM” and “iBT--ROM vs. fH2--fBT--ROM”.
Let
\[
t=\frac{\bar{d}}
       {s_d/\sqrt{n_{\text{sample}}}},
\qquad
\nu=n_{\text{sample}}-1=9,
\]
where
\(n_{\text{sample}}\) is the sample size,
\(\bar{d}\) the sample means,
and \(s_{d}\) the standard deviation of the differences.
The two–sided $p$ value is
\[
p = 2\bigl[1-F_{t,\nu}(|t|)\bigr],
\]
where $F_{t,\nu}(\cdot)$ denotes the cumulative distribution function of the
$t$-distribution with $\nu$ degrees of freedom.
Then we obtain Table~\ref{tab:t-test}.
\begin{table}[htbp]
  \centering
  \caption{Two-sided $t$-test results ($\nu=9$).}
  \label{tab:t-test}
  \begin{tabular}{lcc}
    \toprule
    Comparison & $t$ & $p$ (two-sided) \\ \midrule
    \textbf{rand--ROM} vs. \textbf{fH2--fBT--ROM}        & $16.96$ & $p\ll 10^{-6}$ \\
    \textbf{iBT--ROM} vs. \textbf{fH2--fBT--ROM}   & $3.30$  & $0.0093$ \\ \bottomrule
  \end{tabular}
\end{table}

Table~\ref{tab:t-test} shows
\begin{itemize}
  \item rand--ROM is significantly worse than fH2--fBT--ROM.
  \item For $n_{\text{sample}} = 10$ random seeds, fH2--fBT--ROM (mean $=0.84184$) exceeds iBT--ROM (mean $=0.83765$) by $0.00419$; this difference is statistically significant at the 1\,\% level ($p = 0.0093$).
\end{itemize}
These results support the claim that the proposed fH2--fBT--ROM yields the best performance among the three methods for $r=2$.

%%%%%%%%%%%%%%%%%%%%

%\ifCLASSOPTIONcaptionsoff
 % \newpage
%\fi



\bibliographystyle{IEEEtran}
\bibliography{main.bib}

%\begin{thebibliography}{1}

%\end{thebibliography}



%\vfill


\end{document}